\documentclass[journal]{IEEEtran}

\usepackage{cite}
\usepackage{amsmath}
\usepackage{graphicx}

\usepackage{url}
\usepackage{hyperref}

\usepackage{booktabs}
\usepackage{listings}

\begin{document}

\title{Executable Archaeology: Reanimating the Logic Theorist}

\author{David~Moews and Jeff~Shrager%
\thanks{Authors are in alphabetical order.}
\thanks{David Moews: \texttt{dmoews@fastmail.fm}.}
\thanks{Jeff Shrager is with Bennu Climate, Inc.:
  \texttt{jshrager@gmail.com}.}
\thanks{This work has been submitted to the IEEE for possible publication. Copyright may be transferred without notice, after which this version may no longer be accessible.}
}

\maketitle

\begin{abstract}

The Logic Theorist (LT), created by Allen Newell, J.~C.~Shaw, and
Herbert Simon in 1955--1956, is regarded by some as the first
artificial intelligence program.  It was originally simulated by hand and written in a
pseudocode called the Logic Language (LL), which was later called Information Processing Language One (IPL-I). LT
was revised many times. The first machine-executed version was
written in IPL-II and ran on the JOHNNIAC, and a late version 
was published in 1963 by Einar Stefferud as a pedagogical document and written in IPL-V, the only version of IPL that was formally released to the public.
We describe a reanimation of both the IPL-I (LL) version of LT and the 1963 IPL-V version. Comparing the 
two reanimations illuminates the process of the invention of early AI algorithms and computer languages.

\end{abstract}


\section{Introduction}
\label{sec:intro}

In the summer of 1956, a small group of researchers gathered at
Dartmouth College for a workshop on what John McCarthy had dubbed
``artificial intelligence.'' Among them were Allen Newell and Herbert
Simon, who arrived with a program---albeit one not yet completely running on any 
machine---that could discover proofs of mathematical theorems.
According to Pamela McCorduck's account, it was the only working,
intelligent program presented there~\cite[Ch.~13]{simon1996models},
\cite[p.~123]{mccorduck2004machines}.\footnote{According to
Simon's autobiography \cite[Ch.~13]{simon1996models},
attendee Trenchard More had a program for proving theorems
by natural deduction \cite{more1956,more1957,simon1999}.
It's not clear how developed this was.
}

The Logic Theorist (LT), also called the Logic Theory Machine, would 
ultimately prove 38 of the first 52 theorems in *2 of Alfred North Whitehead 
and Bertrand Russell's \textit{Principia Mathematica}~\cite{LT1957,whitehead1910principia}.\footnote{Star notation follows \textit{Principia}.}  
For Theorem~*2.85, LT discovered a shorter, more direct proof
than the one published in \textit{Principia}. Simon showed the new
proof to Russell himself, who ``responded with
delight''~\cite[p.~167]{mccorduck2004machines}. Edward Feigenbaum,
then an undergraduate taking a graduate course from Simon, recalls the
moment the project's significance was first announced: ``It was just after Christmas vacation---January
1956---when Herb Simon came into the classroom and said, `Over
Christmas Allen Newell and I invented a thinking machine.' And we all
looked blank''~\cite[p.~138]{mccorduck2004machines}.

The Logic Theorist
used heuristic search strategies such as subgoal decomposition
\cite{LT1957}; Newell, Simon, and J.~C.~Shaw suggested that it was a model of
human problem solving~\cite{HPS1958}, and improved the model by adding strategies
that they observed in studies of human theorem proving, such as means-ends analysis \cite{PCT1959}, methods that subsequently influenced artificial intelligence and cognitive models such as the General Problem Solver
(GPS)~\cite{newell1959gps}, SOAR~\cite{laird1991},
and ACT~\cite{anderson1993,anderson1996}.

LT co-evolved with the Information Processing Languages (IPL-I (LL) through IPL-V),
a series of languages created by Newell, Shaw, and Simon  
to extend computational capabilities and understand human thought~\cite[p.~II-17]{cip16june1958}.
The IPLs introduced list manipulation~\cite[section~2.6]{knuth1997} and used dynamic memory allocation~\cite{cmuarchiveshawinterview},
symbols as first-class objects, and recursion.
According to McCarthy \cite{HOPLLISP,mccarthy1960lisp}, the 
list processing of Lisp was
motivated by that of IPL-II, and IPL-V was ancestral
to Joseph
Weizenbaum's SLIP \cite{SLIP1963}, the list-processing package underlying ELIZA~\cite{lane2025eliza}.  Yet while Lisp has thrived for
nearly seven decades, IPL seems to have been almost extinct since the early 1970s.
Although numerous IPL-V implementations existed in the 1960s on
various machines, no maintained implementations appear to be available
in modern environments. The only prior modern attempt known to the
authors is Howard B.~Richman's unpublished IPL-V interpreter, written in
Microsoft BASIC, which he used to run EPAM in the early
1990s.\footnote{Richman wrote to Shrager
(personal communication, Sep~26, 2025, quoted with
permission): ``I just interpreted IPL-V in Microsoft Basic using the
IPL-V manual that I found at the CMU library.''}

The remainder of the present paper describes our
reanimation of two versions of the Logic Theorist:
the original 1956 version written in IPL-I~\cite{LT1956b}, and a  
later 1963 version in IPL-V \cite{stefferud1963lt}.  Following the convention
that the version number tracks the IPL generation, we will call the
1956 IPL-I program \emph{LT1}, the 1957 version
from \cite{LT1957} and \cite{newellshaw1957} \emph{LT2},
and the 1963 IPL-V program \emph{LT5}.

Our work illustrates a methodological approach that might be called
\emph{executable archaeology}: studying old computational systems by
reconstructing and running their original code. Early AIs were not
just proposals but also programs (hand-executed in the case of LT1)
which cannot be completely understood just from their descriptions.
Reanimation gives us a new look into the origins of artificial
intelligence; we return to exactly what this shows in
Section~\ref{sec:contribution}.

In prior work, a team including Jeff Shrager restored
Weizenbaum's original ELIZA to operation on a reconstructed CTSS
running on an emulated IBM 7094~\cite{lane2025eliza}. Although 
LT proves theorems while ELIZA attempts to converse in English,
both are part of the same list-processing lineage coming originally
from IPL.  We reanimated LT5 similarly to ELIZA, by using
an emulated IBM 7094, but for LT1 constructed a new interpreter
for the abstract IPL-I machine; we also tried this approach
with LT5.

\section{Historical Background}
\label{sec:history}

\subsection{The Origins and Versions of the Logic Theorist}

The story of the Logic Theorist begins at the RAND Corporation
in the 1950s, where Simon and Newell wanted to
simulate human problem solving by computer.  Inspired by a 1954 talk
by Oliver Selfridge, Newell started to work on a chess-playing
program, and was joined by Simon and Shaw.  The group discussed
other possible problems as well as chess. In October 1955 they decided to
try to solve geometry problems, but soon switched to
symbolic logic to avoid the problems of simulating diagrams and visual 
perception \cite{laird1991}, \cite[Ch.~13]{simon1996models}.

The version of LT described in the 1956 RAND report
(P-868)~\cite{LT1956a,LT1956b}, and in nearly identical form in the
\textit{IRE Transactions} that September~\cite{LT1956c}, was written in what
was called at the time the ``Logic Language'' (LL).  This was a
pseudocode for an abstract machine which was not implemented at that time. The program
was tested  by human simulation~\cite[Ch.~13]{simon1996models}, 
\cite[p.~xx]{newell1964ipl}. By May
1956, Newell and Simon had noted that the program was ``completely
specified'' and that hand simulations suggested it would prove the
theorems from *2 of \textit{Principia
Mathematica}~\cite[p.~18]{cip1956}.  These simulations famously
employed a computer using ``human components'': Simon's wife, his children, and
graduate students~\cite[Ch.~13]{simon1996models}.

The first version to run on real hardware was written in IPL-II for
the RAND JOHNNIAC~\cite{newellshaw1957},
producing its first machine proof in August
1956~\cite[Ch.~13]{simon1996models}, \cite{newellshaw1957}. The widely cited result
that LT proved 38 of the first 52 theorems in \textit{Principia} comes 
from these experiments~\cite{LT1957}, and the program also discovered
a more elegant proof for Theorem *2.85 than the 
one produced by Whitehead and Russell at around this
time~\cite[Ch.~13]{simon1996models}, \cite{cip8jan1958}.
Newell, Simon, and Shaw proposed LT as a model 
for human problem solving~\cite{HPS1958}
and experimented with human subjects in order to refine
their model~\cite{PCT1959}. This led them to revise
LT, incorporating heuristics such as means-ends
analysis \cite{PCT1959}. Simon and Newell generalized their theory, eventually
creating the General Problem Solver,
GPS~\cite[pp.~19--39]{SHT1959}.

The second version of LT that we reanimated was converted to IPL-V by Fred Tonge, and further developed
by Einar Stefferud as a pedagogical exercise in IPL-V~\cite[Section~I]{stefferud1963lt}.
These  reanimations  bracket the program's
history: LT1 is the 1956 conceptual model as first published, while
LT5 is a 1963 modernized version of that model.

\subsection{IPL: The Invention of List Processing}

\begin{table*}[h]
\centering
\caption{Evolution of the Logic Theorist and IPL.}
\label{tab:lt_versions}
\begin{tabular}{@{}lllll@{}}
\toprule
\textbf{Year} & \textbf{IPL version} & \textbf{LT version} & \textbf{Platform} & \textbf{Remarks} \\ \midrule
1955--1956 & LL (later named IPL-I) & LT1 & \textit{Homo sapiens} & Hand-simulated only. \\
1956--1957? & IPL-II & LT2 & JOHNNIAC & First machine proofs; 38 of 52 \textit{Principia} theorems. \\
1957? & IPL-III & ? & JOHNNIAC & Used for LT but soon abandoned. \\
1957--1959? & IPL-IV & ? & JOHNNIAC & Internal RAND use; transitioning to IPL-V. \\
1958--1970s? & IPL-V & \ & Various & First implemented on the IBM 650 early 1958. \\
1963 & IPL-V & LT5 & \ & Stefferud's pedagogical LT implementation. \\
2026 & IPL-V & LT5 & Simulated IBM 7094 and Python & Reanimation using Stefferud's source code. \\
2026 & IPL-I & LT1 & Python & First known machine execution of LT1. \\ \bottomrule
\end{tabular}
\end{table*}

To implement LT, Newell, Shaw, and Simon needed a programming language
suited to symbolic reasoning. No such language existed, so they
created their ``Information Processing Languages'' (IPLs). 
In the 1956 RAND report (P-868), Newell and Simon call the implementation
language for the Logic Theorist the LL (logic language). It has the
form of an assembly language for an abstract machine whose memory
is divided into working memories, each of which can contain one
element (a variable or connective) of an expression, together with
some annotations, and storage memories, each of which contains a
list of such elements and can be used to hold expressions.  As well
as primitive operation codes, LL can have opcodes defined by LL
routines, so that executing them causes a subroutine call; this
causes a new set of working memories to be used, akin to modern register
windowing. The defining routine of such an opcode can refer to
operands passed in from above. Primitive opcodes can be divided
into subtypes (assign, branch, compare, erase, find, type H (used
for primitive string operations), numerical, put, store, and
test.~\cite{LT1956b}) LL was later retrospectively dubbed
IPL-I~\cite[p.~166]{mccorduck2004machines}.

In IPL-II, the first IPL that ran on a computer, the lists of elements
in IPL-I had to be implemented on the JOHNNIAC, which was done by ordered pairs
of pointers, each pair fitting into one JOHNNIAC word.
As in IPL-I, logical expressions correspond to
lists of elements but now can also include description lists,
lists of properties and values, similar to the later property
lists of Lisp~\cite{newellshaw1957}.  The assembly-language-like system of opcodes
used to implement the Logic Theorist
is very roughly similar to that in LL~\cite{notes1957}.
As in LL, the memory is
divided into working memory and storage, and defined opcodes
can refer to their operands as well as to their own working
memories, but unlike LL, lists can store arbitrary values,
including other lists. IPL-II thus has general list-processing
facilities~\cite{newellshaw1957}.

IPL-III, also on the JOHNNIAC, was used to
revise the Logic Theorist, but was quickly abandoned
as it used too much memory
\cite[p.~xxi]{newell1964ipl}, \cite[p.~II-18]{cip16june1958},
\cite{cip61957}.
IPL-IV ran again on the JOHNNIAC; it was similar to IPL-V,
although its format was more like JOHNNIAC assembly code~\cite{cip16june1958}.
Finally, IPL-V, which first ran on the IBM 650 in early 1958~\cite{DOC1963,iplv1986},
was the version which became the definitive public standard; a manual
was published in 1961~\cite{newell1961ipl}, with a second edition in
1964~\cite{newell1964ipl}.

IPL-V is yet another assembly-style language, running on an abstract machine
organized around \textit{cells}, whose addresses are \textit{symbols}.
Similarly to S-expressions in Lisp, each cell contains an ordered pair of 
symbols.  Some symbols may be used like the atoms of Lisp, but unlike
Lisp, there is no firm distinction between atoms and non-atoms.
The symbol is the basic and central data type of IPL-V
and is used to designate control registers, variables, data structures,
and subroutines.
IPL-V has basic primitives for pushing and popping stacks;
every cell can potentially be pushed and popped, and these stacks are elegantly implemented as IPL-V lists,
or chains of cells.

Control cells
govern execution: H0 is the operand stack, called the
``communication'' cell, H1 is program counter and call
stack, H2 is the free list for dynamic memory allocation, H3 is
the cycle counter, and H5 is a flag bit. Ten working cells (W0--W9)
provide scratch storage. Like Lisp, IPL-V is homoiconic, as both
data and code are lists of cells. 
Routines
are invoked by pushing their symbol (name) onto H1. The IPL machine
interprets the operation indicated in that cell, and then continues to
whatever is the indicated next cell in the list that represents the routine. 
Because H1 is both
program counter and call stack, no special machinery is required to
operate the machine aside from basic stack manipulation instructions.

The language's built-in operations are called J-functions (because
their names begin with ``J'': J0, J1, J2, and so on). There are
approximately 190 of these in the second edition of the
manual, implementing operations from basic list
manipulation to ``generator'' (iterator) protocols. In some implementations
of the original
IPL-V, many J-functions were themselves written in IPL, although some
were implemented at the level of the underlying machine.

\section{Reanimating LT1: The 1956 IPL-I Program}
\label{sec:lt1}

{\em David Moews}

The source code for LT1 was published three times in 1956, in two
versions of paper P-868~\cite{LT1956a,LT1956b} and in the 
\textit{IRE Transactions} paper~\cite{LT1956c}.  The listings are almost identical; I
worked from \cite{LT1956b}, which was the least corrupt.

I transcribed the listing, corrected obvious typos, and
implemented the data structures and operations used from their
descriptions in P-868.  Some opcodes (TCv) were never defined
but obvious, while another (PM) which was used but never defined
seemed to be the same as one (PA) which was defined but never
used, and so on.  Some processes used, as the authors explain
\cite[p.~50]{LT1956b}, were not formally defined, so I simply
made up new opcodes for these; also, some operations had to
be made slightly more precise, and there were some minor bugs
related mostly to lack of coherence between expressions in
storage and their pieces in working memory.  

The first substantial problem was that the routine 
which substituted expressions for variables (LSb)
was trying to substitute a variable with an expression
containing that variable and as a result looping forever.  
Stopping LSb from substituting for non-free variables, and making 
all variables in subproblems non-free, fixed this problem.\footnote{The 
routine matching pairs of expressions, LMc, allows
free variables to occur in both expressions, but doing this correctly 
requires a more complicated method, the unification algorithm.  It's 
often credited to John Alan Robinson (1965)~\cite{robinson1965} but basically
appeared earlier in Jacques Herbrand's doctoral thesis~\cite[Section~18]{lecherbrand2009}.}
I also eventually rewrote pieces of the chaining method (MCh),
owing mostly to various problems coming again from incoherence 
between storage and working memory.

The program in P-868 will state whether a proof was found or not,
but it has no mechanism to print the proof, or even the individual
steps of the proof.  I added new opcodes to print the proof steps
as they were found, and wrote a separate program to reassemble the 
proof, if one existed.  After a few more minor fixes the program seemed 
to be working correctly.  The result of running it is discussed in
Section~\ref{sec:results}.

Since LT1 was not implemented at the time, except by hand
simulation~\cite[Ch.~13]{simon1996models}, \cite[p.~xx]{newell1964ipl}, this is most likely the first time this program has run on non-human hardware.

\section{Reanimating LT5: Stefferud's 1963 IPL-V Program}
\label{sec:lt5}

{\em David Moews}

The source code for LT5 was published and extensively documented
in Stefferud's 1963 report~\cite{stefferud1963lt}.  
Anthony Hay and Shrager transcribed the LT code from Stefferud's
report into a spreadsheet, and Rupert Lane later identified
several transcription errors using his ``gridlock''
tool~\cite{lane2024gridlock}, designed to accurately extract
tabular code from PDF documents.
Shrager wrote a new IPL-V interpreter in Common 
Lisp~\cite{shrager2026,shragerrepo}, but since there were some problems with
it I decided to write a new interpreter in Python.
I worked mostly from the detailed second edition IPL-V
manual~\cite{newell1964ipl}, but also soon realized that
``Simon's J's,'' an old card deck from 1962 saved by the Computer History
Museum~\cite{bochannek2012simonsjs}, could serve as a library for part of the
implementation.

After some debugging and coding, I realized that not
everything was documented in the manual.\footnote{Shrager noticed this as well in his Lisp-based IPL-V interpreter.}
To make Stefferud's code work, I had to copy a hack of  
Shrager's by making a list copy primitive, J73, return a pointer
to an empty cell (with P, Q, SYMB, and LINK all zero) when it was given
a zero (equivalent to NIL in Lisp) input.  Checking this against Rob 
Storey's IBM 7094 emulator \cite{ibsys}, I found that the version of IPL-V there
behaved similarly.  I had been trying to debug my interpreter by comparing
it with IBM 7094 runs, but was handicapped because Stefferud's code
worked very badly on the emulated IBM 7094: the line read primitives
it used didn't exist there, but even after this was worked around,
as I found out later, there was a bug in the M16 and M17 routines
in Stefferud's code \cite{moewsrepo2}.

After some more fixes, I patched the Python interpreter to work
around the bug in Stefferud, as it was making the code crash under the
interpreter.  After doing this and finally writing the line read primitives
(which I hadn't bothered to implement before) the Python interpreter
worked and, when run on Stefferud's code and using the input in Stefferud's
paper, gave results almost identical to the output in his paper.
(There are differences in the effort figures printed out as the
interpreter is not cycle-exact. Also, some numbers printed out
used to index formulae differ as they are raw machine addresses).
After fixing the bug, I was also able to run Stefferud's code using
the IBM 7094 emulator, which also gave very similar results to the
results in his paper \cite{moewsrepo2}.

\section{What the Logic Theorist Proves}
\label{sec:results}

The Logic Theorist attempts to prove theorems in propositional logic
using axioms, definitions, and rules of inference from \textit{Principia Mathematica}, 
*1 and *2; LT5 also adds definitions from *3 and 
*4.  The relevant notation, definitions, and axioms used by \textit{Principia}
are explained in Tables \ref{tab:notation} and \ref{tab:axioms}.
{\em Principia} uses a system of dots for grouping, which we've
replaced by parentheses.
Although definitions *3.01 and *4.01 were used in \cite{stefferud1963lt},
we won't use them here.

The original LT1 tries to prove theorems by {\em substitution}
of expressions into the free variables of an axiom or known
theorem; {\em detachment} (*1.11), or modus ponens, where $q$ is inferred
from $p$ and $p\supset q$; and {\em chaining}, where $p\supset r$
is inferred from $p\supset q$ and $q\supset r$.  Chaining can
be {\em forward} or {\em backward}, depending on which of the
premises is proved immediately by substitution and which becomes
a new subproblem~\cite[pp.~29--45]{LT1956b}.
It also rewrites expressions according to the only definition
it uses, *1.01, but this is done implicitly when one expression
is matched against another and doesn't generate proof 
steps~\cite[p.~34]{LT1956b}.\footnote{\textit{Principia} explains at the beginning of *2 that substitution is used but not given as an explicit rule. Chaining can be justified by theorems *2.05 or *2.06 in \textit{Principia}.}
In LT2, chaining has been split into two methods, forward
and backward~\cite[p.~222]{LT1957}.
In LT5, chaining is again split into two methods,
and definitional rewriting ({\em replacement})
has been split into three explicit methods, one where the whole
expression is rewritten and two where subexpressions 
are~\cite[Section~III]{stefferud1963lt}.
Unfortunately, Stefferud's implementation of forward and
backward chaining is not correct, meaning that LT5 doesn't
always produce valid proofs.

\begin{table}[h]
\centering
\caption{Notation of \textit{Principia}.}
\label{tab:notation}
\begin{tabular}{cl}
\toprule
\textbf{Symbol} & \textbf{Meaning} \\
\midrule
$p=q$\ {\rm Df} & $p$ is defined to be $q$ \\
$p\supset q$ & $p$ implies $q$ \\
$p.q$ & Both $p$ and $q$ \\
$p\equiv q$ & $p$ iff $q$ \\
$p\vee q$ & $p$, $q$, or both \\
$\sim p$ & $p$ is false  \\
\bottomrule
\end{tabular}
\end{table}

Moews tried to replicate the results of \cite{LT1957}
by running LT1 and LT5 on the first 52 theorems (to *2.64)
from {\em Principia}, *2, in 
succession.  The LT5 runs were made with Storey's
IBM 7094 emulator~\cite{ibsys}.  As mentioned above,
to do this, it was
necessary to fix some bugs in Stefferud's code and work 
around functionality not present in the emulator's version
of IPL-V~\cite{moewsrepo2}.  Running with the new Python
IPL-V interpreter gave very similar results.
As in \cite{LT1957}, 
the program is always allowed to use
all preceding theorems, whether it proved them successfully
or not.  

For LT1, a work counter is incremented by 1 whenever it tries
to prove a subproblem by substitution, and when the work counter
exceeds a limit (50) no new subproblems are 
attempted~\cite{moewsrepo}, \cite[p.~47]{LT1956b}.
LT2 had time and space bounds \cite[p.~223]{LT1957},
but the authors don't seem to have stated them.
For LT5, as well as a limit on the number of substitution proofs
attempted (50), there are limits on the number of subproblems
generated (50) and total effort, measured by IPL-V interpreter cycles (200,000).
If counts of these quantities equal or exceed the limits,
no new subproblems are attempted~\cite[pp.~154-155]{stefferud1963lt}.

\begin{table}[h]
\centering
\caption{Axioms and definitions from \textit{Principia}.}
\label{tab:axioms}
\begin{tabular}{ll}
\toprule
\textbf{Name} & \textbf{Axiom/Def}\\
\midrule
*1.01  & $p\supset q \, = \,(\sim p) \vee q\ {\rm Df}$\\
*3.01  & 
$p.q \, = \, \sim((\sim p)\vee (\sim q))\ {\rm Df}$\\
*4.01  & $p\equiv q\, = \, (p\supset q).(q\supset p)\ {\rm Df}$\\
*1.2   & $(p\vee p)\supset p$\\
*1.3   & $q\supset(p\vee q)$\\
*1.4   & $(p\vee q)\supset(q \vee p)$\\
*1.5   & $(p\vee(q\vee r))\supset (q\vee(p \vee r))$\\
*1.6   & $(q\supset r)\supset((p\vee q)\supset(p\vee r))$\\
\bottomrule
\end{tabular}
\end{table}

Table~\ref{tab:results} shows the results of LT1, LT2, and LT5
attempting to prove the first 52 theorems in *2. 
LT5 outputs a proof for *2.13, although Newell, Shaw, and Simon
explain~\cite[p.~222]{LT1957} that it's impossible for Logic 
Theorist to prove this theorem.  This is because Stefferud uses
an incorrect forward chaining rule: he infers $q C s$ from
$q C r$ and $r\supset s$, where $C$ is a connective.  This 
is not a valid rule in general as it fails if $C$ is $\equiv$
but works here where $C$ is $\vee$
(we infer *2.13, $p \vee \sim \sim \sim p$, from 
*2.11, $p \vee \sim p$
and a substitution instance of *2.12,
$(\sim p) \supset \sim \sim \sim p$).
Also, LT5 finds a dubious proof of *2.2, where
$(c\vee p)\supset(p\vee q)$ is inferred from the axiom *1.4 via 
substitution.  In LT5, $c$ is a free variable,
so this could be construed as valid in the sense that you can
substitute for $c$ so as to make the proof step and the overall
proof work.  In general though, this may indicate another problem with LT5,
similar to the problems with matching and substitution found with LT1.

LT1 and LT5 usually found the same or very similar proofs;
there were differences when one used forward and the other
backward chaining, or in one using theorems written with $\vee$ as premises
and the other using equivalent theorems which were written with $\supset$.

\def\Y{$+$}
\def\N{$-$}

\begin{table}[h]
\centering
\caption{Theorems attempted.}
\label{tab:results}
\begin{tabular}{@{} lll @{\hspace{3pt}} l @{\hspace{3pt}} l @{}}
\toprule
\textbf{Name} & \textbf{Formula} & \textbf{LT1} & \textbf{LT2}\cite{LT1957} & \textbf{LT5} \\
\midrule
*2.01 & $(p \supset \sim p) \supset \sim p $ & \Y & \Y \cite[p.~219]{LT1957} & \Y \\
*2.02 & $q \supset (p \supset q) $ & \Y & \Y \cite[p.~224]{LT1957} & \Y \\
*2.03 & $(p \supset \sim q) \supset (q \supset \sim p) $ & \Y & \ & \Y \\
*2.04 & $(p \supset (q \supset r)) \supset (q \supset (p \supset r)) $ & \Y & \ & \Y \\
*2.05 & $(q \supset r) \supset ((p \supset q) \supset (p \supset r)) $ & \Y& \ & \Y \\
*2.06 & $(p \supset q) \supset ((q \supset r) \supset (p \supset r)) $ & \Y& \ & \Y \\
*2.07 & $p \supset (p \vee p) $ & \Y & \Y \cite[p.~226]{LT1957} & \Y \\
*2.08 & $p \supset p $ & \Y & \ & \Y \\
*2.1 & $ (\sim p) \vee p $ & \Y & \ & \Y \\
*2.11 & $p \vee \sim p $ & \Y & \ & \Y \\
*2.12 & $p \supset \sim \sim p $ & \Y & \ & \Y \\
*2.13 & $p \vee \sim \sim \sim p $ & \N & \N \cite[p.~222]{LT1957} & ? \\
*2.14 & $ (\sim \sim p) \supset p $ & \Y & \ & \Y \\
*2.15 & $( (\sim p )\supset q) \supset ( (\sim q) \supset p) $ & \N & \ & \N \\
*2.16 & $(p \supset q) \supset ( (\sim q) \supset \sim p) $ & \N & \ & \N \\
*2.17 & $( (\sim q) \supset \sim p) \supset (p \supset q) $ & \N & \Y \cite[p.~228]{LT1957} & \N \\
*2.18 & $( (\sim p )\supset p) \supset p $ & \Y & \ & \Y \\
*2.2 & $p \supset (p \vee q) $ & \Y & \ & ? \\
*2.21 & $ (\sim p) \supset (p \supset q) $ & \Y & \ & \Y \\
*2.24 & $p \supset ( (\sim p) \supset q) $ & \Y & \ & \Y \\
*2.25 & $p \vee ((p \vee q) \supset q) $ & \Y & \ & \N \\
*2.26 & $ (\sim p) \vee ((p \supset q) \supset q) $ & \Y & \ & \Y \\
*2.27 & $p \supset ((p \supset q) \supset q) $ & \Y & \ & \Y \\
*2.3 & $(p \vee (q \vee r)) \supset (p \vee (r \vee q)) $ & \Y & \ & \Y \\
*2.31 & $(p \vee (q \vee r)) \supset ((p \vee q) \vee r) $ & \N & \N \cite[p.~220]{LT1957} & \N \\
*2.32 & $((p \vee q) \vee r) \supset (p \vee (q \vee r)) $ & \N & \ & \N \\
*2.36 & $(q \supset r) \supset ((p \vee q) \supset (r \vee p)) $ & \N & \ & \N \\
*2.37 & $(q \supset r) \supset ((q \vee p) \supset (p \vee r)) $ & \N & \ & \N \\
*2.38 & $(q \supset r) \supset ((q \vee p) \supset (r \vee p)) $ & \Y & \ & \N \\
*2.4 & $(p \vee (p \vee q)) \supset (p \vee q) $ & \N & \ & \N \\
*2.41 & $(q \vee (p \vee q)) \supset (p \vee q) $ & \N & \ & \N \\
*2.42 & $(( \sim p) \vee (p \supset q)) \supset (p \supset q) $ & \Y & \ & \N \\
*2.43 & $(p \supset (p \supset q)) \supset (p \supset q) $ & \Y & \ & \Y \\
*2.45 & $ (\sim (p \vee q)) \supset \sim p $ & \Y & \Y \cite[p.~219]{LT1957} & \Y \\
*2.46 & $ (\sim (p \vee q)) \supset \sim q $ & \Y & \ & \Y \\
*2.47 & $ (\sim (p \vee q)) \supset ( (\sim p) \vee q) $ & \Y & \ & \Y \\
*2.48 & $ (\sim (p \vee q)) \supset (p \vee (\sim q)) $ & \Y & \Y \cite[p.~229]{LT1957} & \Y \\
*2.49 & $ (\sim (p \vee q)) \supset ( (\sim p) \vee (\sim q)) $ & \Y & \ & \Y \\
*2.5 & $ (\sim (p \supset q)) \supset ( (\sim p) \supset q) $ & \Y & \ & \Y \\
*2.51 & $ (\sim (p \supset q)) \supset (p \supset \sim q) $ & \Y & \ & \Y \\
*2.52 & $ (\sim (p \supset q)) \supset ( (\sim p) \supset \sim q)$ & \Y & \ & \Y \\
*2.521 & $ (\sim (p \supset q)) \supset (q \supset p) $ & \Y & \ & \Y \\
*2.53 & $(p \vee q) \supset ( (\sim p) \supset q) $ & \Y & \ & \N \\
*2.54 & $( (\sim p )\supset q) \supset (p \vee q) $ & \Y & \ & \N \\
*2.55 & $ (\sim p) \supset ((p \vee q) \supset q) $ & \Y & \ & \Y \\
*2.56 & $ (\sim q) \supset ((p \vee q) \supset p) $ & \Y & \ & \N \\
*2.6 & $( (\sim p) \supset q) \supset ((p \supset q) \supset q) $ & \N & \ & \N \\
*2.61 & $(p \supset q) \supset (( (\sim p) \supset q) \supset q)$ & \Y & \ & \Y \\
*2.62 & $(p \vee q) \supset ((p \supset q) \supset q) $ & \Y & \ & \Y \\
*2.621 & $(p \supset q) \supset ((p \vee q) \supset q) $ & \Y & \ & \Y \\
*2.63 & $(p \vee q) \supset (( (\sim p) \vee q) \supset q) $ & \Y & \ & \Y \\
*2.64 & $(p \vee q) \supset ((p \vee (\sim q)) \supset p) $ & \N & \ & \N \\
\ & \ & \ & \ & \ \\
\ & Number of correct proofs: & 40 & 38 & 33+$2\cdot ?$ \\
\bottomrule
\end{tabular}
\begin{center} \Y{} means that a correct proof was found while \N{} means that no
proof was found.  For ?, see the text.
\end{center}
\end{table}

An exception to this is Theorem *2.55.  We show the proof output by LT5
for this theorem in Figure
\ref{proof5255}.\footnote{Although the effort limit was set to 200,000, a bug in the system makes it display as 000000.  Also, in the emulated IBM 7094 run, the output lines were cut off on the right-hand side, and we have reconstructed them here.}
The proof begins from Theorem~*2.25, which is the minor premise
of detachment; the major premise is a substitution instance of
*2.53.  Applying detachment then gives *2.55.  The proof steps
output by LT1 form a slightly different proof, Figure \ref{proof1255}.
In this case, we start by proving a result definitionally equivalent to *2.53, and
use it as the minor premise in detachment, the major premise being a substitution
instance of *1.5.  After definitional rewriting, the result is again *2.55.

\begin{figure*}
\begin{lstlisting}
TO PROVE
*2.55   -PI((PVQ)IQ)
           1.   -PI((-(PVQ))VQ)                   , SUBLEVEL REPLACEMENT
           12312   -PI((-(PVQ))VQ)                , SUBLEVEL REPLACEMENT.  REJECTED PROBLEM
           2.   --PV((PVQ)IQ)                *1.01, REPLACEMENT
           3.   PV((PVQ)IQ)                  *2.53, DETACHMENT
 PROOF FOUND.
          GIVEN                         *2.25     AV((AVB)IB)
          SUBSTITUTION                  3.        PV((PVQ)IQ)
          GIVEN                         *2.53     (AVB)I(-AIB)
          DETACHMENT                    *2.55     -PI((PVQ)IQ)
          Q.E.D.
EFFORT              LIMIT 000000        ACTUAL 28443
SUBPROBLEMS         LIMIT 50            ACTUAL 3
SUBSTITUTIONS       LIMIT 50            ACTUAL 4
\end{lstlisting}
\caption{Proof of *2.55 from LT5. \lstinline|I| denotes $\supset$, \lstinline|V| denotes
$\vee$, and \lstinline|-| denotes $\sim$.}
\label{proof5255}
\end{figure*}

\begin{figure*}
\begin{lstlisting}
PROOF_DET: successfully proved ( ~ p -> ( ( p \/ q ) -> q ) ) by detachment 
  theorem ( ( A \/ ( B \/ C ) ) -> ( B \/ ( A \/ C ) ) ) 
  remaining-problem ( ~ ( p \/ q ) \/ ( ~ ~ p \/ q ) )
PROOF_SUB: successfully proved ( ~ ( p \/ q ) \/ ( ~ ~ p \/ q ) ) by substitution 
  theorem ( ( A \/ B ) -> ( ~ A -> B ) )
\end{lstlisting}
\caption{Proof steps of *2.55 from LT1. \lstinline|\\/| denotes $\vee$ and \lstinline|->| denotes $\supset$.}
\label{proof1255}
\end{figure*}

\section{What Executable Archaeology Contributes}
\label{sec:contribution}

It is reasonable to ask what is learned by running this old code that
could not be learned by reading it, or by reading the abundant
secondary literature on the Logic Theorist. Here are some
possible answers.

\textbf{It draws attention to gaps and differences.}
The most striking single finding of the LT1 reanimation is an
absence: the published 1956 program can find proof steps and 
recognize that a proof exists, but cannot assemble
or output the proof itself.  Although this is obvious, turning
the code into a working program forces you to attend to it. 
It illustrates what Newell and Simon
thought was and wasn't important to formalize at this point.

LT1 and LT5 are similar programs but
differ by about an order of magnitude in size---roughly 300
IPL-I instructions against around 3,000 lines of
IPL-V.  Part of this difference is because LT5 has more
functionality, for example, being able to
assemble and print proofs; another partial reason is that it uses
a general-purpose language, IPL-V, instead of LL, which was
specialized for the LT; a third is that LT1 leaves the details
of memory management more or less unspecified, while in LT5
cells are explicitly requested and freed.  Reimplementing 
both languages places them on a common footing and permits a 
better comparison of LT1 and LT5.

\textbf{It shows that some semantics survive only in artifacts, not
in documentation.}  The behavior of J73 when called with NIL
doesn't seem to be documented; it was
only discoverable by trying to run Stefferud's code, or by
comparing with the version of IPL-V on the IBM 7094 emulator.
Specifications and manuals don't cover every detail. Reanimation, on the other hand,
requires that such details be clarified.

\textbf{It points out errors in the published record.}  The 1956 
IPL-I source listings contain various inconsistencies, typos, and
bugs, and Stefferud's 1963 code also turned out to be wrong and
buggy in various ways.  Repairing these problems, and the requirement
that every repair be justified and recorded, is part of the ``archaeology''
in executable archaeology.

Most importantly, \textbf{reanimation makes historical claims checkable.} The 1956
report~\cite{cip1956} claimed, on the basis of hand simulation, that the Logic
Theorist would prove the theorems of \textit{Principia} *2; the 
1957 paper~\cite{LT1957} reported the famous 38-of-52 result for the JOHNNIAC version.
Until now, such claims rested on testimony and on archival output
listings.  The reanimations allow these types of claims to be tested, 
and for many further experiments to be made.

\section{Related Work}
\label{sec:related}

Most directly related to this work
is the ELIZA reanimation by Lane
et al.~\cite{lane2025eliza}, in which a team including Shrager
restored Weizenbaum's original ELIZA to operation on a
reconstructed CTSS running on an emulated IBM~7094. That project
reconstructed the complete original hardware and operating system
stack. Apart from this, the present work also takes a different, yet parallel approach, emulating 
abstract machine specifications in modern languages.

Reimplementing the Logic Theorist's \textit{algorithms} in a modern
language is a relatively straightforward exercise---essentially an
introductory AI course project, and one that has surely been done many
times. A 1968 paper, for example, describes an implementation of LT in
LISP~1.5~\cite{millstein1968logic}. The present work is a
fundamentally different undertaking: rather than rewriting LT's logic
in a modern language, it constructs emulators of the IPL-I
and IPL-V abstract machines and runs the original code, transcribed
from the 1956 and 1963 reports. The distinction is
significant: a Lisp reimplementation demonstrates that LT's algorithms
can be expressed in Lisp (which is unsurprising), while the present
work demonstrates that the original IPL code, as written and published
by Newell, Simon, and Shaw's team, functions as described---and,
moreover, enables experiments on the real, working systems of machine
and program.

\section{Next Steps}
\label{sec:next}

The IPL-V interpreters developed for this project are not specific to
the Logic Theorist. They implement the general IPL-V abstract machine
and can, in principle, run any IPL-V program. Work is in progress on
a corpus of early IPL-V AI programs. The most significant targets are
the General Problem Solver (GPS)~\cite{newell1959gps} and EPAM
(Elementary Perceiver and Memorizer)~\cite{feigenbaum1961epam}. 
EPAM source code is available in the Feigenbaum archives at Stanford, and
archival work to locate complete source code for GPS is in progress.
Now that proven IPL interpreters exist, running these programs should
require substantially less effort than was required for LT.

The IPL-I interpreter, repaired LT1 source, and proof-assembly tools
are open sourced at \cite{moewsrepo}.
Shrager's IPL-V interpreter in Common Lisp is open sourced at
\cite{shragerrepo}, and Moews's IPL-V interpreter
in Python, and directions for running LT5 under Storey's IBM 7094
emulator, are open sourced at \cite{moewsrepo2}.

\section*{A Personal Coda}

{\em Jeff Shrager}

I studied under both Newell and Simon at Carnegie
Mellon University in the 1980s, where my work focused on modeling the
cognition of scientists and engineers. IPL and LT were long since
history by that time, and neither came up in classes or
conversation. 
There is a
certain closure in now studying Newell, Simon, and Shaw \textit{as}
scientist-engineers: reverse-engineering their design decisions,
reconstructing their reasoning, and trying to understand how they
invented AI and cognitive science. The subject of that training and
its object have folded into one another.

\section*{Acknowledgments}

The following people (listed in no particular order) contributed to
the LT5 effort in ways ranging from code transcription to weekly
encouragement: Anthony Hay, Arthur Schwarz, Leigh Klotz, David
M.\ Berry, Rupert Lane, Amy Majczyk, Emily Davis, Ethan Ableman, and
Paul McJones. 

We thank the Computer History Museum for preserving and making
available ``Simon's J's,'' the original 1962 IPL-V J-function card
deck \cite{bochannek2012simonsjs},
which was useful in various ways. We also thank Alexander
Magoun and David Walden for editorial guidance, including the
suggestion to combine Moews's reanimation efforts, together
with Shrager's previous account \cite{shrager2026}, which portions
of this paper are based on, into this single account.

Finally, we acknowledge Allen Newell, Herbert Simon, and Cliff Shaw,
whose work this paper merely recreates.

{\footnotesize Portions of the LT5 debugging effort, and portions of
the drafting of this paper, were carried out with the aid of various
LLMs, but all concepts and final editorial decisions were the
authors' alone.}

\end{document}